\documentclass[11pt]{article}

\usepackage[preprint]{acl}
\usepackage{booktabs}
\usepackage{times}
\usepackage{latexsym}
\usepackage{booktabs}
\usepackage{tabularx}

\usepackage[T1]{fontenc}

\usepackage[utf8]{inputenc}

\usepackage{microtype}

\usepackage{inconsolata}

\usepackage{graphicx}
\usepackage{microtype}
\usepackage{hyperref}
\usepackage{url}
\usepackage{booktabs}
\usepackage{algorithm}
\usepackage{algpseudocode}
\usepackage{amsmath}

\usepackage{graphicx}
\usepackage{booktabs}
\usepackage{multirow}
\usepackage{subcaption} 

\usepackage[capitalise]{cleveref}
\makeatletter
\AddToHook{cmd/appendix/before}{\def\cref@section@alias{appendix}}
\makeatother

\crefname{appendix}{Appendix}{Appendices}
\Crefname{appendix}{Appendix}{Appendices}
\crefname{subappendix}{Appendix}{Appendices}
\Crefname{subappendix}{Appendix}{Appendices}
\crefname{subsubappendix}{Appendix}{Appendices}
\Crefname{subsubappendix}{Appendix}{Appendices}

\let\oldappendix\appendix
\renewcommand{\appendix}{%
  \oldappendix%
  \crefalias{section}{appendix}%
  \crefalias{subsection}{subappendix}%
  \crefalias{subsubsection}{subsubappendix}%
}
\usepackage{xspace}
\newcommand{\method}[0]{\textsc{CreativeInstruct}\xspace}

\newcommand{\llamathreeone}{LLaMA-3.1 8B\xspace}
\newcommand{\qwentwofiveseven}{Qwen2.5 7B\xspace}
\newcommand{\qwenthreeeight}{Qwen3 8B\xspace}
\newcommand{\qwentwofivethirtytwo}{Qwen2.5 32B\xspace}
\newcommand{\qwenthreethirtytwo}{Qwen3 32B\xspace}

\title{\method: Scalably Teaching LLMs to Balance \\ Quality, Creativity, and Diversity}
\author{Ananya Sahu$^1$ \quad Mohit Bansal$^2$ \quad Elias Stengel-Eskin$^3$
\\
\\
$^1$Columbia\quad
$^2$UNC Chapel Hill \quad
$^3$University of Texas at Austin
}

\begin{document}
\maketitle

\begin{abstract}
While post-training improves the capabilities of large language models (LLMs), it generally lowers their output diversity and creativity, negatively impacting tasks that explicitly require creativity (e.g., story generation) as well as those that require it implicitly, e.g., reinforcement learning (RL). We instead propose \method{}, a scalable instruction-tuning method that teaches LLMs to
balance creative, base-model-like generations with the quality of
post-trained models, by learning to inject special
{\tt{[StartCreativity]}} spans that bias generation toward creativity.
Furthermore, we introduce a structural diversity metric based on graph edit distance, which captures narrative-level variation missed by purely lexical and semantic metrics. 
On narrative generation, \method{} matches or exceeds the diversity of both multi-model baselines and distilled variants of their outputs, without sacrificing quality or requiring multiple models at inference time. These results are mirrored in our human evaluation, where we find that annotators rate \method generations as more creative than the post-trained LLMs' generations in 70.3\% of cases. 
We also show the benefits of creative models as a substrate for RL: GRPO applied to a \method{} checkpoint improves by $\sim4\%$ on AMC and $\sim5\%$ points on MATH over the same training applied to the post-trained checkpoint.\footnote{Code: \url{https://github.com/ananya-sahu/CreativeInstruct}.}

\end{abstract}

\section{Introduction}
\begin{figure}[ht]
    \centering
    \includegraphics[width=\columnwidth]{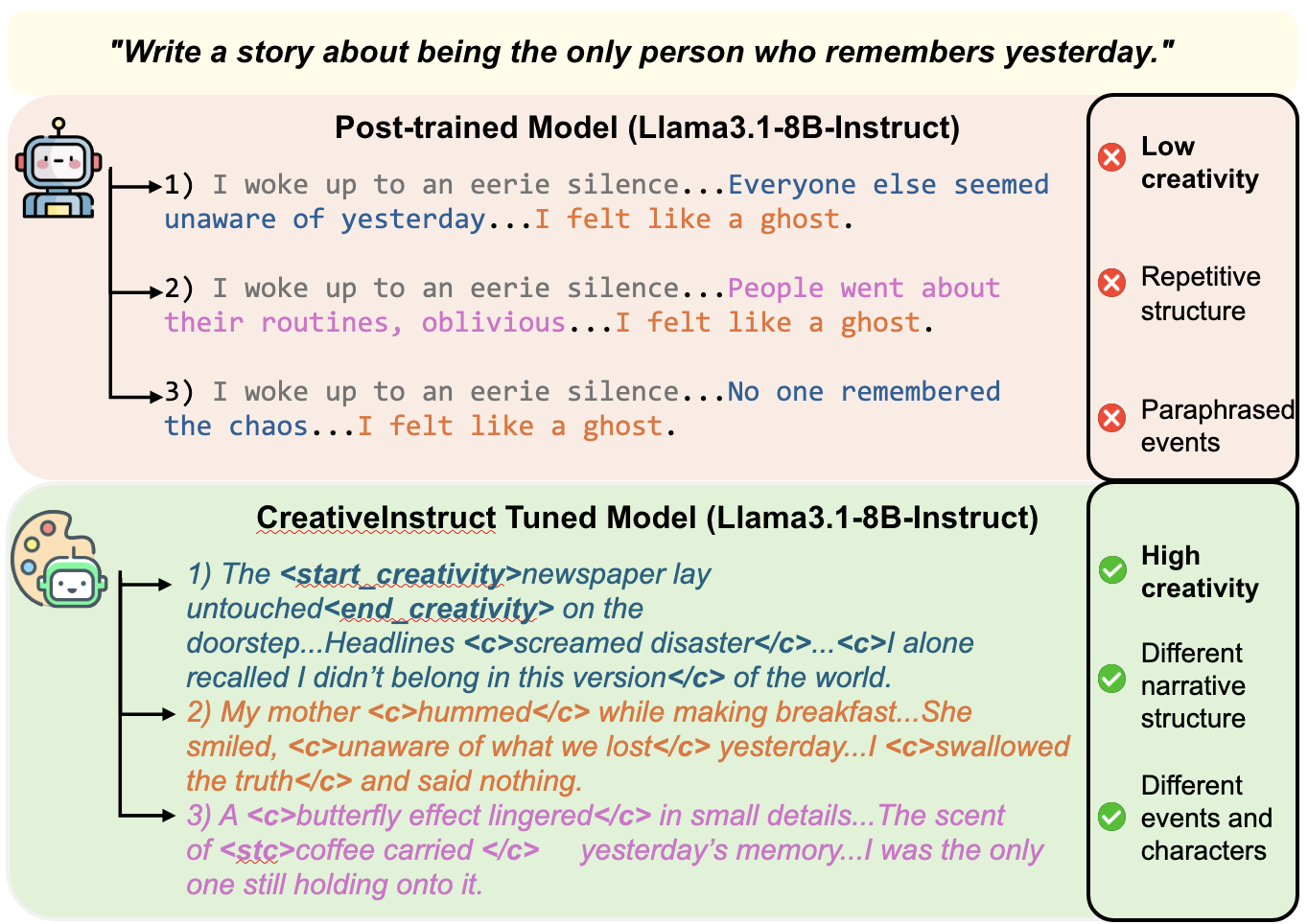}
    \caption{Instruction-tuned models succeed in following instructions but produce non-creative, repetitive outputs. \method{} maintains the aligned model's quality while improving its creativity by learning to insert creativity-triggering tokens.
    }
    \label{fig:teaser}
\end{figure}

Creativity is a critical aspect of human intelligence, characterized by unpredictability and diversity \cite{macedo2002assessing,boden1995creativity}.
Both of these qualities are essential for generating novel ideas and solving open-ended problems \cite{friedman2016increase}. 
While post-training greatly improves the abilities of large language models (LLMs) to follow instructions, reason, and produce safe outputs \citep{lu2025alignment}, it generally comes at a cost to the diversity and creativity of LLMs \citep{west2025base}.  
Across different model families, outputs from post-trained -- i.e., aligned -- LLMs tend to exhibit homogeneity \cite{wenger2025we}. 
Moreover, scaling alone does not mitigate this loss of diversity: regardless of model size, aligned model outputs tend to converge into homogeneous clusters \cite{jiang2025artificial,chih2026cog}. 
The consequences of reduced diversity apply both to tasks that explicitly call for diversity and ones that require it implicitly.
In generative settings like creative writing, the loss of diversity leads to repetitive outputs that fail to explore the breadth of plausible generations for a given prompt \cite{zhang2025verbalized}. 
More broadly, generation diversity is a critical factor across a range of tasks, including reinforcement learning (RL), where diverse rollouts are essential for effective exploration and policy improvement \cite{hu2025diversity, chih2026cog}.

To improve creativity in post-trained models, 
recent work has introduced decoding frameworks that incorporate outputs from both a base model -- which has greater diversity -- and an aligned variant -- which has greater instruction-following ability \cite{wang2025optimizing, feng2025don}. 
However, these methods incur latency and computational overhead at test-time, requiring two models. 
Moreover, these frameworks require access to base models, which are not always released. 

To scalably balance diversity with quality, we introduce \method{}, 
an instruction-tuning method that trains a single, unified LLM to steer its own output towards higher creativity.
As illustrated in \cref{fig:method}, \method{} leverages multi-model inference-time approaches to create instruction-tuning data for training a unified model. 
We first identify a subset of general-purpose instruction-tuning data pertaining to writing (as opposed to math, code generation, etc.) and use BACo -- a competitive inference-time routing strategy \citep{wang2025optimizing} that routes between base and aligned models based on uncertainty and semantic roles of tokens -- to generate responses.\footnote{We note that \method{} is agnostic to the routing method chosen.}
We track base model spans in the generated outputs and surround them with two
special tokens, {\tt{[StartCreativity]}} and {\tt{[EndCreativity]}}. We then
fine-tune post-trained models on this data so they learn when to switch
between the base model's diversity and creativity and the aligned model's
instruction-following. 
Critically, at test time, the model self-injects creativity tokens, deciding when to switch. This allows it to increase diversity without compromising quality by automatically selecting key points where switching is warranted. 

To evaluate \method{} and other baselines beyond surface-level diversity, we propose an LLM-based graph edit distance metric, which we call LLM Graph-Edit-Distance (GED). 
LLM-GED operates on a set of outputs; it extracts abstract narrative graphs for each story and compares the graphs to each other, producing pairwise graph edit distance scores. 
These scores measure narrative-level diversity: LLM-GED uses abstract narrative units rather than lexical ones, identifying low-diversity outputs such as the aligned model's outputs in \cref{fig:teaser}, which follow the same pattern. 

Using LLM-GED and other standard diversity metrics, we evaluate \method{} on narrative generation tasks across 5 models ranging from 7B to 32B. \method{} consistently provides gains over both trained and test-time baselines; for example, \llamathreeone trained with \method{} yields approximately 48\% relative gains in semantic diversity and 63\% gains in structural diversity over the base instruct model. 
Human evaluation confirms these results with outputs from \method{} rated as more creative than those generated by aligned models in 70.3\% of comparisons.
Moreover, despite the fact that \method{}'s training data is generated by running inference-time routing -- in this case, BACo \citep{wang2025optimizing} -- on training examples, \method{} consistently outperforms BACo at test-time, across model families and sizes, with average gains of 29\% in semantic diversity and 28\% in structural diversity. 
This indicates that \method{} can generalize beyond the training data, allowing it to replace routing via test-time heuristics with a learned, internal switch. 

We further show that fine-tuning with our creativity tags yields higher
quality scores than fine-tuning without them, while maintaining comparable
or higher diversity scores.
Because \method{} relies on general-purpose instruction-tuning data, it is scalable. We demonstrate that performance continues to improve as we scale up the training data.
Finally, we demonstrate that the benefits of creativity instruction-tuning extend to RL. 
Specifically, we compare the performance of standard post-trained \qwenthreeeight checkpoint, 
finetuned via GRPO \citep{shao2024deepseekmath} on math data, 
to the performance of an identical GRPO finetuning process applied to a \method{} checkpoint. 
We find that the GRPO applied to the \method{} checkpoint achieves higher performance, with a $~4\%$ gain on AMC \cite{yang2024qwen2} after being trained on MATH \cite{hendrycks2021measuring} data. 
This highlights the importance of creativity not only in writing tasks but as an LLM objective more broadly. 

\section{Related Work}
\paragraph{Enhancing Diversity in Post-Trained Models.}
The trade-off between model alignment and diversity presents a fundamental challenge in creative generation tasks, which require both. 
Prior work has shown that post-training can reduce diversity and creativity \citep{kirk2024understanding,spangher2025creative, west2025base}. 
Several training-time approaches have been proposed to address this, including training against diversity objectives \citep{ismayilzada2025creative, chung2503modifying,lanchantin2025diverse}.
\citet{chung2025literarytaste} curate and incorporate literary preference datasets into training for human preference. 
More recently, \citet{samuel2026recovering} use base and instruct models to create preference pairs to optimize for diversity.
Unlike these methods, which treat diversity as an explicit training objective at the level of a whole output, we fine-tune on creativity-tagged, diverse sub-spans, letting the model learn where to automatically inject creativity localized by span.

Test-time frameworks have also been explored as approaches for improving diversity and creativity \citep{zhang2025verbalized}, 
including methods that route between base and aligned models \citep{wang2025optimizing, feng2025don} or between models from different families \citep{liu2026no}. 
However, these methods require both models to be loaded simultaneously, incurring additional memory overhead and latency.
In our work, we leverage BACo \cite{wang2025optimizing}, one representative dual-model decoding framework, to curate training data for a single unified model. 
Interestingly, we find that \method{} outperforms BACo at test-time, despite the latter using multiple models; 
this suggests that learning from the routing signal goes beyond distillation, and allows for greater generalization than test-time routing, while also incurring a lower test-time cost.

\paragraph{Prior Work on Diversity Optimization for RL.}
Recent work has demonstrated that diversity in model generations plays a critical role in problem-space exploration during reinforcement learning \citep{chih2026cog, bahlousboldi2026vector}. 
In particular, \citet{yao2026diversity} show that optimizing learning policies for diversity enables them to discover more diverse and robust solutions for mathematical reasoning tasks. 
Rather than incorporating an explicit diversity-based reward during reinforcement learning, we begin with a \method{}-trained model that already exhibits increased diversity, and train it with standard GRPO objectives to improve mathematical reasoning performance.

\section{Methodology}
\subsection{\method{}}
\paragraph{Background.}We draw on BACo \cite{wang2025optimizing} as a data generation framework, which is an inference-time token-level routing framework that combines token outputs from a base LLM with its aligned model counterpart, and leads to increased diversity while maintaining quality in generative tasks.
BACo uses routing strategies based on uncertainty and semantic roles at each token to optimize generations for diversity and maintain quality. While we leverage BACo, in our work, it is important to note that any multi-model routing framework can be used to generate our data following our data curation methodology -- BACo was chosen based on its performance. 

\begin{figure*}
    \centering
    \includegraphics[width=\linewidth]{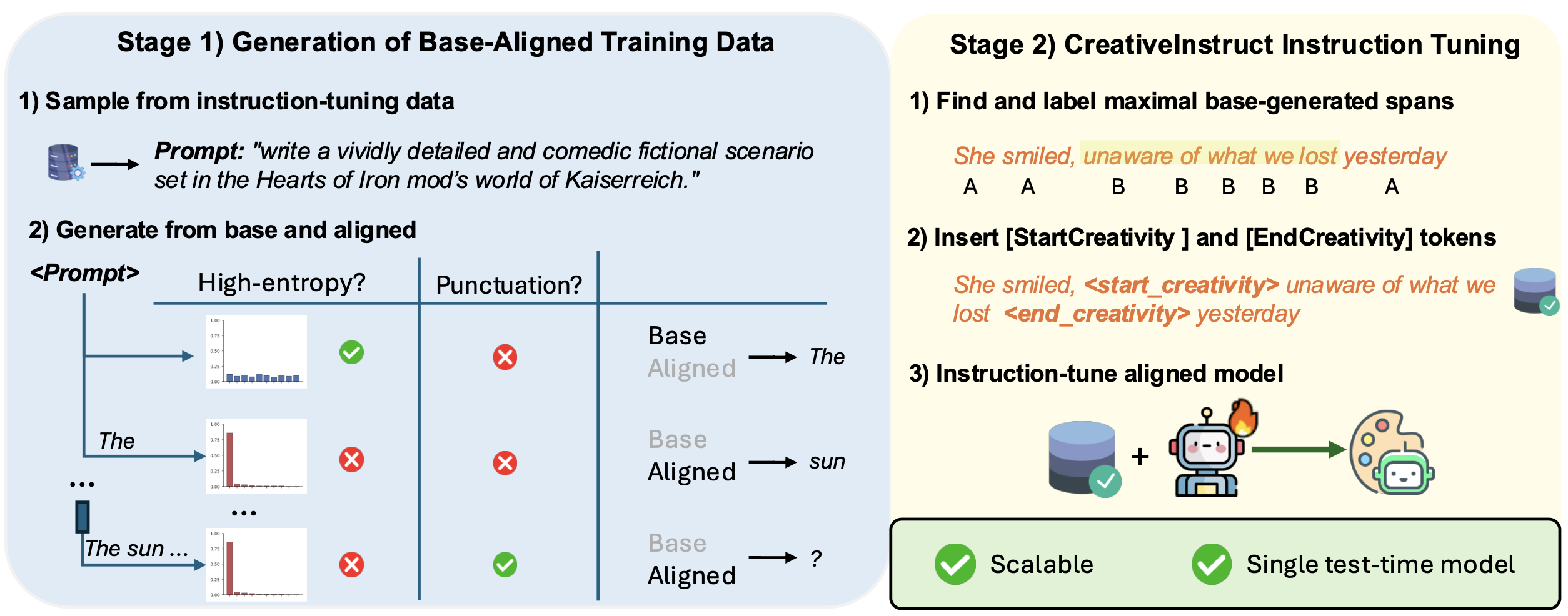}
    \caption{
    \method{} operates by first creating a dataset of creativity-tagged instruction-tuning data. For each prompt, token generation is routed between base and aligned model variants based on token entropy and whether the token represents a sentence boundary, following \citet{wang2025optimizing}. 
    Contiguous spans of tokens generated from the base model are then marked with new {\tt{[Start-]}} and {\tt{[EndCreativity]}} tokens, and the aligned model is trained on this data. At test time, the new \method{} model generates its own start- and end-creativity tokens, emulating both base and aligned models in a single checkpoint.}
    \label{fig:method}
\end{figure*}

We use BACo prob+punc, the best-performing routing variant on diversity metrics, to generate our training data. The BACo prob+punc variant routes based on token probability and token type.  Punctuation and formatting tokens are routed to the aligned model to preserve coherent sentence formatting and grammatical structure while all other tokens are routed based on token entropy: high-entropy tokens are routed to the base model to encourage generation diversity, and lower-entropy tokens are routed to the aligned model.
This routing heuristic is illustrated in \cref{fig:method} (Stage 1).

\paragraph{Data and Dataset Construction.}For our main results, we draw prompts from a subset of the T\"{u}lu V3 SFT dataset \cite{lambert2024tulu3}, filtered for English writing-related prompts, comprising 4,000 unique prompts. 
We generate 3 outputs per prompt, yielding 12,000 training samples. 
All models are trained on T\"{u}lu data. 
As depicted in \cref{fig:method}, for each prompt we track which model each token in the final generated output comes from (base or aligned) and place creativity token markers around spans of tokens that come from the base model. 
We also place markers around spans of aligned and base model tokens that have the same probability (within a delta of 0.005). 
We generate all data using the base and instruct variant of each model family and size.
The only exception to this is \qwenthreethirtytwo variant which does not have a publicly-released base variant. Here we train on \qwentwofivethirtytwo generated data, demonstrating that transfer between models is possible when base models are unavailable.  
\paragraph{Training and Inference.}We fine-tune aligned models on our generated dataset with creativity tokens injected, using LoRA~\cite{hu2021lora}. We evaluate all methods on the Narrative Discourse dataset \cite{tian2024large}. Additional training parameter details can be found in ~\cref{appendix:training_setup}.

\subsection{LLM Graph Edit Distance Metric}
To measure narrative diversity, we measure the structural diversity between stories using an LLM-guided graph-based abstraction that captures entities, events, and temporal structure. We refer to this score as the LLM Graph Edit Distance (LLM-GED) score. 
LLM-GED scaffolds an LLM judge's reasoning about diversity by framing the problem as one of measuring the distance between graphs.
Specifically, the judge is instructed to first represent each story as an abstract event graph \(G = (V, E)\), where nodes correspond to entities (e.g., characters, locations, objects) and events, and edges capture semantic relations and temporal ordering. The full prompting schema is provided in \cref{appendix:ged_prompt}.

LLM-GED aims to measure true narrative diversity rather than lexical differences. 
Therefore, we canonicalize entities into abstract identifiers (e.g., \textit{Character1}, \textit{Location1}). 
We further prompt the model to use semantic roles \citep{fillmore2003form} when canonicalizing entities and events, such as \textit{agent}, \textit{affected}, and \textit{causes}, along with temporal ordering through directed \textit{next\_event} relations. 

Given two stories, the judge estimates a normalized graph edit distance corresponding to the minimum number of structural edit operations -- node and edge insertions, deletions, or relabelings -- required to transform one abstract event graph into another. 
We normalize the score to account for story length differences:
\[
\text{nGED}(G_A, G_B) =
\frac{\text{GED}(G_A, G_B)}{\max(|G_A|, |G_B|)},
\]
where \(|G|\) denotes the total number of nodes and edges in the abstract graph representation.

We compute all pairwise structural distances jointly in a single prompt. Given a set of \(N\) stories, the LLM is instructed to reason over all stories under a shared canonical schema and produce a full \(N \times N\) matrix of pairwise normalized GED estimates. 
The final LLM-GED score we report is the average normalized GED across all story pairs. 
We further discuss this LLM-based metric and compare it to a pipeline that deterministically computes pairwise edit distances between canonicalized graphs in \cref{append:ged_details}; while the two methods produce comparable results, the pipeline is far more expensive, so we opt for the LLM-based version. 
In \cref{append:llm_ged}, we additionally find LLM GED correlates better than semantic metrics on structural diversity judgments, shown in \cref{tab:correlations_diversity}.

\section{Experiments and Results}

\begin{table*}[t]
\centering
\small
\setlength{\tabcolsep}{3.2pt}
\renewcommand{\arraystretch}{1.1}

\begin{tabular}{llcccccc}
\toprule
\textbf{Model Family}
& \textbf{Method}
& Cos-D (M) $\uparrow$
& Cos-D (Q) $\uparrow$
& Sem Ent $\uparrow$
& Vendi $\uparrow$
& NLI Div $\downarrow$
& LLM-GED $\uparrow$ \\
\midrule

\multirow{4}{*}{\textsc{Llama-3.1 (8B)}}
& BACo      & 0.255 & 0.254 & 2.112 & 3.499 & 0.065 & 0.374 \\
& Instruct  & 0.309 & 0.284 & 2.126 & 2.913 & 0.040 & 0.366 \\
& CrPO   & 0.350 & 0.359 & \textbf{2.127} & 3.852 & \textbf{0.028} &0.514  \\
& Distill   & 0.444 & 0.362 & 2.063 & 3.975 & 0.052 & 0.523 \\
& \method{} & \textbf{0.458} & \textbf{0.417} & 2.038 & \textbf{4.749} & 0.085 & \textbf{0.545} \\

\midrule

\multirow{4}{*}{\textsc{Qwen2.5 (7B)}}
& BACo      & 0.348 & 0.313 & 2.264 & 3.204 & 0.025 & 0.476 \\
& Instruct  & 0.290 & 0.249 & \textbf{2.286} & 2.681 & \textbf{0.023} & 0.442 \\
& Distill   & 0.335 & 0.295 & 2.263 & 3.298 & 0.029 & 0.456 \\
& \method{} & \textbf{0.395} & \textbf{0.380} & 2.177 & \textbf{4.167} & 0.037 & \textbf{0.523} \\

\midrule

\multirow{4}{*}{\textsc{Qwen2.5 (32B)}}
& BACo      & 0.259 & 0.238 & 2.012 & 2.573 & 0.054 & 0.370 \\
& Instruct  & 0.250 & 0.200 & 2.234 & 2.360 & \textbf{0.024} & 0.416 \\
& Distill   & 0.281 & 0.275 & 2.123 & 2.852 & 0.050 & 0.396 \\
& \method{} & \textbf{0.295} & \textbf{0.282} & \textbf{2.894} & \textbf{3.358} & 0.052 & \textbf{0.483} \\

\midrule

\multirow{4}{*}{\textsc{Qwen3 (8B)}}
& BACo      & 0.265 & 0.245 & 2.023 & 2.974 & 0.085 & 0.395 \\
& Instruct  & \textbf{0.399} & 0.344 & 1.451 & 3.257 & 0.130 & 0.453 \\
& Distill   & 0.306 & 0.292 & \textbf{2.189} & 2.983 & 0.051 & 0.475 \\
& \method{} & 0.369 & \textbf{0.351} & 2.078 & \textbf{3.926} & \textbf{0.038} & \textbf{0.498} \\

\midrule

\multirow{3}{*}{\textsc{Qwen3 (32B)}}
& Instruct  & 0.253 & 0.206 & \textbf{2.280} & 2.515 & \textbf{0.030} & 0.466 \\
& Distill   & \textbf{0.391} & \textbf{0.376} & 1.784 & \textbf{3.484} & 0.067 & 0.376 \\
& \method{} & 0.301 & 0.288 & 1.909 & 3.290 & 0.059 & \textbf{0.478} \\

\bottomrule
\end{tabular}
\caption{Semantic diversity metrics and LLM-GED across model families. $\uparrow$ indicates higher is better, $\downarrow$ indicates lower is better. Bold indicates best value per column within each metric and model family. }
\label{tab:unified_diversity_metrics}
\end{table*}

\paragraph{Baselines and Models.}We compare models fine-tuned using \method to post-trained Instruct models, as well as to BACo \citep{wang2025optimizing} applied at test-time. 
Note that the latter baseline requires twice the test-time compute budget of the others. We additionally include a Distill baseline, which uses the same corpora generated with BACo but omits the {\tt{[Start-]}} and {\tt{[EndCreativity]}} tags. This baseline directly measures the contribution of these explicit markers. 
Lastly for the \llamathreeone
variant, we compare to one additional baseline, \llamathreeone trained with the creative preference optimization (CrPO) algorithm \cite{ismayilzada2025creative}. 
We use \llamathreeone \cite{grattafiori2024llama}, \qwentwofiveseven  \cite{Yang2024Qwen25TR}, and Qwen 3 8B \cite{yang2025qwen3}. 
We also test our methods on larger models: \qwentwofivethirtytwo, and \qwenthreethirtytwo. 
We evaluate BACo applied to each model, other than \qwenthreethirtytwo, for which there is no base model released (further motivating our method, which does not rely on having test-time base-model access).

\begin{table*}[t]
\centering
\small
\renewcommand{\arraystretch}{1.15}
\begin{tabular}{llcccc}
\toprule
\textbf{Model Family} & \textbf{Method} & \textbf{Coherence} & \textbf{Fluency} & \textbf{Relevance} & \textbf{WQRM} \\
\midrule

\multirow{4}{*}{\textsc{Llama-3.1 (8B)}}
& BACo      & 3.31 & \textbf{4.30} & 3.41 & 5.73 \\
& Instruct  & \textbf{3.69} & 4.09 & \textbf{3.99} & 5.93 \\
& CrPO       & 3.45 & 4.02 & 3.09 & 6.09 \\
& Distill   & 3.43 & 3.98 & 3.42 & 5.85 \\
& \method{} & 3.55 & 4.22 & 3.39 & \textbf{6.65} \\

\midrule

\multirow{4}{*}{\textsc{Qwen2.5 (7B)}}
& BACo      & 4.10 & 4.35 & 3.82 & \textbf{6.47} \\
& Instruct  & \textbf{4.18} & \textbf{4.62} & \textbf{4.26} & 6.17 \\
& Distill   & 2.98 & 3.72 & 2.76 & 5.81 \\
& \method{} & 3.82 & 4.02 & 3.84 & 6.42 \\

\midrule

\multirow{4}{*}{\textsc{Qwen2.5 (32B)}}
& BACo      & 4.35 & 4.30 & 4.69 & 6.17 \\
& Instruct  & \textbf{4.54} & 4.44 & \textbf{4.91} & 5.84 \\
& Distill   & 3.75 & 4.02 & 3.85 & 5.84 \\
& \method{} & 3.89 & \textbf{4.67} & 4.31 & \textbf{6.34} \\

\midrule

\multirow{4}{*}{\textsc{Qwen3 (8B)}}
& BACo      & \textbf{3.64} & \textbf{4.26} & \textbf{3.68} & 6.86 \\
& Instruct  & 3.50 & 3.74 & 3.36 & 6.56 \\
& Distill   & 2.38 & 2.45 & 2.29 & 6.32 \\
& \method{} & 3.47 & 3.84 & 3.02 & \textbf{6.90} \\

\midrule

\multirow{3}{*}{\textsc{Qwen3 (32B)}}
& Instruct  & \textbf{4.45} & \textbf{4.85} & \textbf{4.76} & \textbf{7.39} \\
& Distill   & 3.74 & 3.74 & 4.17 & 5.76 \\
& \method{} & 4.36 & 4.14 & 4.01 & 6.32 \\

\bottomrule
\end{tabular}
\caption{Quality evaluations (Likert 1--5) and Writing Quality Reward Model (WQRM; 1--10) for narrative generation across model families.}
\label{tab:quality_main}
\end{table*}

\paragraph{Metrics.}For evaluation, we follow the diversity metrics used in \citet{wang2025optimizing} as well as our LLM-GED score. The full set of diversity metrics includes commonly used semantic metrics: embedding based cosine dissimilarity (all-MiniLM and Qwen-3 embeddings), semantic entropy \citep{kuhn2023semantic}, Vendi Score (SimCSE) \citep{friedman2022vendi}, and NLI Diversity \citep{stasaski2022semantic}; 
and narrative structure diversity measured using our LLM-based Graph Edit Distance (GED) metric. We also evaluate on lexical metrics included in \cref{appen:lexical}. Taken together, these metrics enable comprehensive evaluation of diversity in creative text generation.
In addition, we evaluate our generations for quality along dimensions of coherence, fluency, and relevance following the evaluation framework in \citep{liu2023g}. All LLM based metrics use GPT5-mini.\footnote{We use gpt-5-mini-2025-08-07 variant of GPT5-mini} We also use a writing quality reward model (WQRM) informed by human preferences from \citet{chakrabarty2025ai} to evaluate quality.

\subsection{Main Results}

\Cref{tab:unified_diversity_metrics}
show the semantic and graph based diversity scores, and quality scores of narrative generations.
\method{} consistently improves semantic and structural diversity across most model families, often surpassing both Instruct and BACo baselines.
On \llamathreeone{}, \method{} improves over Instruct and BACo on the majority of
semantic metrics, including a +0.149 gain in MiniLM cosine dissimilarity over Instruct and +0.203 over BACo respectively. On LLM-GED we see similar trends with \method{} achieving 17 point gains over BACo and Instruct.
These gains hold
across model families and scales: \method{} achieves the highest mean performance across the majority of metrics over both Instruct and BACo
for \qwentwofiveseven{} and \qwenthreeeight{}. 
Similarly
\qwentwofivethirtytwo{} improves over Instruct on most metrics,
with a notable gain of +0.082 in Qwen Cos-D (Cos-D (Q)).

Compared to the Distill baseline, \method{} consistently achieves stronger semantic and structural diversity across multiple model families, with higher LLM-GED in every setting and substantial improvements in cosine-based diversity and Vendi Score for most models.
Moreover, for \llamathreeone, \qwentwofivethirtytwo, and \qwenthreeeight, these diversity gains are achieved while also obtaining the highest Writing Quality Reward Model (WQRM) score within each model family.
The exception is \qwenthreethirtytwo, where Distill does better across most automatic diversity metrics, 
although \method{} continues to produce more structurally diverse generations according to the LLM-GED metric.
We additionally report lexical diversity metrics in ~\cref{tab:lexical_diversity_grouped}, which show similar trends, with \method{} consistently outperforming both the standard instruction-tuned baseline and BACo across smaller model families.

\cref{tab:quality_main} reports automatic quality evaluations, showing that \method{} generally maintains competitive performance on coherence, fluency, relevance, and the Writing Quality Reward Model (WQRM). The Distill variant generally underperforms \method{} on automatic quality metrics, indicating that, in addition to generally improving diversity, adding {\tt{Start-}} and {\tt{EndCreativity}} tags help preserve generation quality during instruction tuning.

\begin{table*}[t]
\centering
\small
\renewcommand{\arraystretch}{1.15}
\begin{tabular}{p{0.47\textwidth} p{0.47\textwidth}}
\hline
\textbf{\method{}} & \textbf{\textsc{Instruct}} \\
\hline
\textit{``As I walked down the street, I noticed something was off. [...] He looked at me confused and said, `What tornado? I don't remember anything out of the ordinary happening yesterday.' ''}
&
\textit{``I woke up to an eerie silence. Everyone around me seemed normal, yet I knew something was off. [...] I recalled the sunburn I got at the beach, the argument with my best friend, and the stunning sunset.''}
\\
\hline
\textit{``As the day began, I noticed a small crowd gathering, each person whispering to each other. [...] But then, I saw it -- a newspaper lying on the ground, yesterday's date still visible. [...] I was the only one who remembered yesterday.''}
&
\textit{``I woke up to an eerie silence. Yesterday was still etched in my mind like a vivid movie. The world outside was stuck in a perpetual loop, reliving the same mundane moments. [...] I felt like a ghost, doomed to relive the same day forever.''}
\\
\hline
\textit{``I don't know what's happening, but it's as if everyone else has forgotten yesterday. [...]People around me are going about their daily routines like
nothing unusual happened, but I know better. [...] The world is moving on, but I'm stuck in the past.''}
&
\textit{``I woke up to an eerie silence, the streets empty and still. [...] I recalled the storm that ravaged the city, the sirens wailing, and the people fleeing for their lives. [...] I felt like a lone sentinel, the only one who remembered the horrors of the past.''}
\\
\hline
\end{tabular}
\caption{Qualitative comparison on prompt \emph{``Write a story about being the only person who remembers yesterday''} between \llamathreeone-\method and \llamathreeone instruct model. While \method generations vary in framing and narrative structure, the instruct generations repeatedly rely on nearly identical openings (e.g., \emph{``I woke up to an eerie silence''}) and recurring catastrophic motifs. See \cref{tab:qualitative_full_generations} for full stories.
\label{tab:qualitative_sample_generations}}
\end{table*}

\begin{table}[t]
\centering
\small
\setlength{\tabcolsep}{4pt}
\renewcommand{\arraystretch}{1.1}
\begin{tabular}{lcc}
\toprule
\textbf{Variant} &
\shortstack{\textbf{Per Prompt}\\\textbf{Group}} &
\shortstack{\textbf{Corpus}\\\textbf{Wide}} \\
\midrule
Instruct  & 18.125 & 11.994 \\
Distill   & 26.567 & 16.767 \\
BACo      & 17.927 & 12.116 \\
\method{} & \textbf{37.129} & \textbf{24.659} \\
\bottomrule
\end{tabular}
\caption{Proper noun uniqueness (\%) at the prompt-group and corpus-wide levels for each training variant. Higher values indicate less repetition of proper nouns (e.g., character or place names) across generations. \method{} significantly outperforms all other variants in both settings (Mann-Whitney U test on per-prompt scores, $p < 0.001$).}
\label{tab:proper_noun_uniqueness}
\end{table}

\paragraph{\method{} Improves Entity Diversity.}\Cref{tab:qualitative_sample_generations} illustrates a pattern of formulaic repetition in baseline generations, with similar entities appearing repeatedly. \citet{yao2025understanding} show LLMs can fall into repetitive generation patterns, and that repetition features are often associated with named entities.
To quantify this, we compute proper noun uniqueness: the ratio of unique
proper nouns (e.g., character or place names) to total proper nouns,
identified using the spaCy NER tagger \citep{honnibal2020spacy}.
We measure this at two
granularities: the prompt-group level (averaged across all generations
sharing the same prompt) and the corpus-wide level, on generations from
\method{}, Instruct, BACo, and Distill, all using \llamathreeone{}.  
\Cref{tab:proper_noun_uniqueness} shows \method{} achieves the highest proper
noun uniqueness at both granularities, 37.1\% vs.\ 26.6\% for the strongest
baseline (Distill) at the prompt-group level, and 24.7\% vs.\ 16.8\%
corpus-wide, more than doubling Instruct's uniqueness at both granularities
(18.1\% and 12.0\%, respectively). 
This indicates that \method{}'s diversity gains extend to reduced repetition of names and entities across generations.

\begin{figure}[t]
    \centering
    \includegraphics[width=\columnwidth]{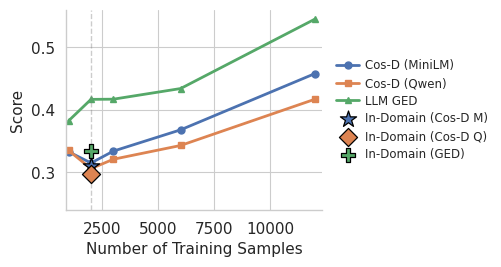}
    \caption{
    Scaling synthetic training data improves semantic and narrative diversity for \llamathreeone. Distilled corresponds to training on synthetic corpus from same prompts in evaluation.
    }
    
    \label{fig:scaling_diversity}
\end{figure}

\paragraph{Effects of Scaling on Creativity of Narrative Generations.}
\Cref{fig:scaling_diversity} shows Cosine Dissimilarity and LLM-GED metrics as a function of the number of T\"{u}lu SFT samples. 
In general, increasing the number of training samples increases diversity scores; \llamathreeone{} achieves its best performance at 12000 samples. 
Additionally, the scores have not yet plateaued at the highest number of samples, indicating that further scaling the training data may yield continued diversity gains. 

\paragraph{Importance of Data Diversity.}To evaluate the role of the \emph{type} of data we train on, we additionally evaluate an ``in-domain'' variant.
Here, we train \method{} on Narrative Generation data rather than T\"{u}lu data, testing to what degree data domain matters; this data does not scale since there is a very fixed amount of it. 
We generate 10 outputs per training prompt, resulting in a total of 2,020 training samples.
Overall, we see that general-purpose instruction-tuning data from T\"{u}lu leads to higher average scores across both metrics than the in-domain model even at the same data set size. 
This indicates the importance of data diversity in enabling models to learn creative generalizations.

\subsection{Human Evaluation}
\citet{jaarsveld2012creative} define creativity to be a composite of divergent and convergent thinking, a paradigm reflecting diversity and quality of LLM generation outputs. 
To further evaluate the creativity of our instruction-tuned model, we conduct a human evaluation with three annotators. 
Annotators consist of non-author undergraduate and graduate students with backgrounds in NLP and related technical fields. 
We first generate prompts using GPT-5 for 50 diverse topics, each designed to elicit a five-sentence story. For each prompt and each system (Instruct, \method{}) we sample 10 generations from the system and randomly select five.  
Annotators are shown all 5 generations from both systems, side-by-side, one prompt at a time; an example of three generations can be seen in \cref{tab:qualitative_sample_generations}.  
The full annotation interface and instructions can be found in \cref{append:annotator_inst} and \cref{append:annotator_interface}. 
System identities are anonymized and the presentation order of the systems is randomized. 
Annotators are asked to provide pairwise preference judgments comparing the two system generations along three dimensions: diversity, quality, and overall creativity. 
To assess inter-annotator agreement, a subset of 14 prompts are evaluated by all three annotators.  

\paragraph{Results.}We compute inter-annotator agreement using Cohen's $\kappa$ on pairwise judgments. For creativity we see high agreement ($\kappa$ = 0.720), and we see moderate agreement for diversity ($\kappa$ = 0.417). 
We attribute this gap to the multidimensional nature of the diversity judgment, which leaves more room for annotators to weight different axes of diversity (e.g., lexical, structural, semantic) differently, relative to more unified -- albeit more subjective -- creativity judgments as defined by our annotation guidelines (see \cref{append:annotator_inst}).  
However, we observe low agreement for quality ($\kappa =  -0.167$). This may be due to the small size of our annotations; we instead rely on automatic metrics for quality evaluations, omitting quality from our human comparisons.
\Cref{tab:llama_ft_vs_baseline} shows the human evaluation results comparing generations from the \method{} model and the Instruct baseline. 
\method{} achieves 70.3\% preference for creativity over baseline instruct. 
This suggests that the gains seen in \cref{tab:unified_diversity_metrics} and \cref{tab:quality_main} are also detected by human annotators.

\begin{table}[ht]
\centering
\renewcommand{\arraystretch}{1.3}
\resizebox{\linewidth}{!}{
\begin{tabular}{lcc}
\hline
\textbf{System} & \textbf{Diversity} & \textbf{Creativity} \\
\hline
\textsc{\method} & 57.4\% & $70.3\%^{*}$ \\
\textsc{Instruct} & 42.6\% & $29.7\%^{*}$ \\
\hline
\end{tabular}
}
\caption{Win-rate across evaluation dimensions for \llamathreeone fine-tuned vs.\ baseline. ($n = 50$); $^{*}$ indicates significance (two-sided binomial test).
}
\label{tab:llama_ft_vs_baseline}
\end{table}

\subsection{Creativity for RL}
Successful reinforcement learning (RL) requires a balance between exploration and exploitation \citep{sutton1998reinforcement}. 
Current LLM RL post-training methods like GRPO \citep{shao2024deepseekmath} sample roll-outs from a policy, which are then scored according to a reward; a lack of exploration -- i.e., a lack of diversity -- in these rollouts can hamper learning \citep{chih2026cog}, as the policy effectively only explores a single solution or strategy. 
We explore whether the improved diversity \method{} offers can provide a better starting-point for RL by training \qwenthreeeight models with GRPO.
We train two models, a baseline \qwenthreeeight model and the same model fine-tuned with \method. 
We train both models using GRPO loss, for the same number of steps, 1000, with Huggingface's TRL library \citep{vonwerra2020trl}. 
During training we generate 8 rollouts with context length 2048; other hyperparameters are given in \cref{append:rl-train}.
We train on a 12k training split of MATH \cite{hendrycks2021measuring}  and evaluate on both in-domain test data from MATH \cite{hendrycks2021measuring} and on AMC \cite{yang2024qwen2}. Our results are averaged over three runs with different random seeds.

\begin{table}[t]
\centering
\resizebox{\linewidth}{!}{
\begin{tabular}{lcc}
\toprule
\textbf{Model} 
& \textbf{MATH} 
& \textbf{AMC} \\
\midrule
Instruct baseline
& 0.374
& 0.432\\
Instruct + RL
& 0.409
& 0.438\\
\method baseline
& 0.424
& 0.428\\

\method + RL
& \textbf{0.459} 
& \textbf{0.478}\\

\bottomrule
\end{tabular}
}
\caption{
Accuracy comparison on math reasoning benchmarks after training with RL with \qwenthreeeight-\method and \qwenthreeeight Instruct models. }
\label{tab:transposed_accuracy}
\end{table}

\paragraph{Results.}\Cref{tab:transposed_accuracy} shows the results of RL-training on \method{} and the Instruct model. 
Our more diverse model has an advantage in downstream math reasoning tasks. 
Appendix \cref{fig:ood_difficulty} breaks these accuracies down by difficulty on AMC. 
Overall, these results suggest that increasing diversity during instruction tuning not only preserves but can improve downstream reasoning performance, with consistent gains across most difficulty levels. 
First, \cref{tab:transposed_accuracy} shows that, while training on the additional T\"{u}lu instruction-tuning data improves MATH500 performance, it does not improve AMC performance: while the MATH performance improves by $3.5\%$ between the Instruct baseline and the \method{} baseline, the AM performance actually drops slightly.
After RL, the picture changes for the OOD data, with the \method{}+RL model outperforming the Instruct+RL model by 4\% on AMC. 
This indicates that, while \method{} does not consistently improve a model's math abilities alone, its improved diversity creates a better substrate for RL, enabling better OOD generalization.

\section{Conclusion}
We introduce \method{}, an instruction-tuning approach for improving diversity and creativity in language model generation while maintaining competitive output quality. 
Our method trains on data generated by routing between base and aligned models to obtain creative spans, enabling a single model to recover diversity without relying on multi-model inference-time decoding. 
We further introduce a structural diversity metric based on LLM-assisted graph edit distance to better capture narrative-level variation beyond lexical and embedding-based measures. 
This correlates better than semantic based metrics for structural diversity judgements. 
Experiments across model families on the Narrative Discourse dataset show consistent improvements in diversity metrics over baseline variants with minimal impact on quality. 
Human evaluations indicate a preference for our model in terms of creativity. Finally, we observe that models trained with \method also improve downstream performance in reinforcement learning settings on mathematical reasoning tasks.

\section*{Acknowledgments}

We would like to thank Hanqi Xiao, Atin Pothiraj, and Nithin Sivakumaran for their annotation effort.

\bibliography{custom}

\appendix

\section{Methodological Details}

\label{appendix:training_setup}
\subsection{Fine-Tuning and Inference Parameter Details}
We use LoRA fine-tuning for our \method{} trained models. We train with rank $r=32$, $\alpha=64$, dropout $0.05$, applied to all attention and MLP projection layers. We train for up to $10$ epochs with early stopping (patience$=3$), using a batch size of $8$, learning rate of $2\times10^{-4}$, and a cosine scheduler with warmup ratio $0.03$. At test-time, we generate outputs with the same decoding parameters as BACo: temperature $1.0$, top-$p$ $0.9$, and a maximum of $512$ new tokens.

\begin{table}[h]
\centering
\begin{tabular}{ll}
\toprule
\textbf{Hyperparameter} & \textbf{Value} \\
\midrule
Per-device batch size & 4 \\
Gradient accumulation steps & 4 \\
Effective batch size & 16 \\
Learning rate & $1\times10^{-6}$ \\
LR scheduler & constant \\
Max grad norm & $0.1$ \\
Number of rollouts & $8$ \\
Max completion length & $2048$ \\
Max epochs & $10$ \\
Early stopping patience & $500$ steps \\
Precision & bfloat16 \\
\bottomrule
\end{tabular}
\caption{Hyperparameters for RL training with GRPO.}
\label{append:rl-train}
\end{table}

\section{Additional Results and Discussion}
\label{sec:append_additional}

\begin{table*}[t]
\centering
\small
\setlength{\tabcolsep}{4pt}
\begin{tabular}{p{0.22\linewidth} p{0.22\linewidth} p{0.08\linewidth} p{0.10\linewidth} p{0.30\linewidth}}
\toprule
\textbf{S1} & \textbf{S2} & \textbf{\# Edits} & \textbf{LLM-GED} & \textbf{Edit Explanation} \\
\midrule

He entered the room. He saw the body. He ran away.
&
He walked into the chamber. He noticed the corpse. He fled.
&
0
&
0.00
&
Pure lexical variation; all events and temporal structure are equivalent after canonicalization. \\

\midrule

Alice woke up. She had breakfast. Then she left.
&
Alice woke up. She left the house. She then had breakfast.
&
4
&
0.45
&
Temporal reordering of events: delete edges (Wake $\rightarrow$ Breakfast, Breakfast $\rightarrow$ Leave) and insert (Wake $\rightarrow$ Leave, Leave $\rightarrow$ Breakfast). \\

\midrule

He opened the box. He found a key. He left.
&
He opened the box. He found a key. He examined it. He left.
&
3
&
0.30
&
Insertion of event node (Examine) with corresponding edges (Find $\rightarrow$ Examine, Examine $\rightarrow$ Leave). \\

\midrule

The guard opened the gate. He closed the gate.
&
The guard opened the gate. He locked the gate.
&
1
&
0.10
&
Single event relabel: (Close $\rightarrow$ Lock); all other nodes and edges unchanged. \\

\midrule

The dog chased the cat.
&
The cat chased the dog.
&
2
&
0.25
&
Role reversal: relabel agent and affected edges, swapping subject-object structure. \\

\midrule

He sat down. He opened his laptop. He started working.
&
He sat down. He started working.
&
2
&
0.28
&
Deletion of event node (Open laptop) and corresponding temporal edge adjustments. \\
\bottomrule
\end{tabular}
\caption{Examples illustrating normalized graph edit distance (LLM-GED) across different types of variation. LLM-GED assigns zero distance to paraphrases (row 1) while increasing with temporal, structural, and semantic changes.}
\label{tab:nged_examples}
\end{table*}

\subsection{Graph Edit Distance Metric Automatic Evaluation}
\label{append:llm_ged}

We validate the effectiveness of our LLM-GED metric at capturing different levels of narrative diversity in four settings for 100 paired stories originating from the same prompt.  
\begin{enumerate}
    \item \textbf{Same Stories (A,A)}: pair the same story generated as a control
    \item \textbf{Lexical Shifts (A,A')}: replace 50 percent of words with their synonyms 
    \item \textbf{Temporal Shifts (A,A')}: Switch the order of events that occur in the story from end to beginning. 
    \item \textbf{Different Stories (A,B)}: pick a story belonging to a different prompt 
\end{enumerate} 
We hypothesize an ideal metric for measuring narrative diversity will rank same stories as the least diverse, lexical shifts as slightly more diverse, temporal shifts as more diverse and different stories entirely as most diverse. In \cref{tab:correlations_diversity} we compare our metric's correlations with narrative shift rankings against lexical and semantic metrics from \cite{wang2025optimizing} and show our metric correlates the highest with ground truth rankings. 

\begin{table}[t]
\centering
\small
\setlength{\tabcolsep}{4pt}
\renewcommand{\arraystretch}{1.1}

\begin{tabularx}{\columnwidth}{lXc}
\toprule
\textbf{Category} & \textbf{Metric} & \textbf{Correlation} \\
\midrule

Semantic & Cosine Dissimilarity (all-MiniLM) & 0.852 \\
Semantic & Cosine Dissimilarity (Qwen) & 0.698 \\
Semantic & Semantic Entropy & -0.410 \\
Semantic & Vendi Score (SimCSE) & 0.654 \\
Semantic & NLI Diversity & 0.687 \\
\midrule

Narrative Structure & LLM-GED & \textbf{0.889} \\
\midrule

Lexical & Dist-1 & 0.231 \\
Lexical & Dist-2 & 0.505 \\
Lexical & Dist-3 & 0.511 \\
Lexical & EAD-1 & 0.505 \\
Lexical & EAD-2 & 0.522 \\
Lexical & EAD-3 & 0.522 \\
Lexical & Self-BLEU & 0.511 \\
Lexical & Self-ROUGE-L & 0.678 \\
Lexical & Vendi Score (N-Gram) & 0.505 \\
\bottomrule
\end{tabularx}

\caption{Correlation between diversity metrics and a reference ranking of increasingly diverse narrative variations.}
\label{tab:correlations_diversity}
\end{table}

\begin{table*}[t]
\centering
\scriptsize
\setlength{\tabcolsep}{2.2pt}
\renewcommand{\arraystretch}{0.95}

\resizebox{\textwidth}{!}{
\begin{tabular}{llccccccccc}
\toprule
\textbf{Model Family}
& \textbf{Method}
& Dist-1 
& Dist-2 
& Dist-3 
& EAD-1 
& EAD-2 
& EAD-3 
& Self-BLEU 
& Self-ROUGE-L 
& Vendi (NGram) \\
\midrule

\multirow{3}{*}{\textsc{\llamathreeone}}
& BACo
& 0.320 & 0.700 & 0.858 & 0.339 & 0.696 & 0.844 & \textbf{0.539} & \textbf{0.224} & 6.227 \\

& Instruct
& 0.261 & 0.668 & 0.881 & 0.294 & 0.695 & 0.889 & 0.536 & 0.169 & 4.911 \\

& \method{}
& \textbf{0.366} & \textbf{0.793} & \textbf{0.944} & \textbf{0.421} & \textbf{0.821} & \textbf{0.946} & 0.387 & 0.150 & \textbf{6.596} \\

\midrule

\multirow{3}{*}{\textsc{\qwentwofiveseven}}
& BACo
& 0.248 & 0.697 & \textbf{0.914} & 0.274 & 0.719 & 0.919 & 0.516 & \textbf{0.157} & 5.354 \\

& Instruct
& 0.252 & 0.690 & 0.908 & 0.286 & 0.730 & \textbf{0.925} & \textbf{0.551} & 0.166 & 4.901 \\

& \method{}
& \textbf{0.305} & \textbf{0.715} & 0.890 & \textbf{0.349} & \textbf{0.757} & 0.909 & 0.476 & 0.166 & \textbf{6.311} \\

\midrule

\multirow{3}{*}{\textsc{\qwentwofivethirtytwo}}
& BACo
& 0.249 & 0.686 & 0.898 & 0.278 & 0.712 & 0.908 & 0.542 & 0.175 & 5.123 \\

& Instruct
& 0.247 & \textbf{0.693} & \textbf{0.911} & 0.275 & \textbf{0.726} & \textbf{0.927} & 0.559 & 0.176 & 4.673 \\

& \method{}
& \textbf{0.288} & 0.677 & 0.855 & \textbf{0.324} & 0.707 & 0.862 & \textbf{0.562} & \textbf{0.204} & \textbf{5.814} \\

\midrule

\multirow{3}{*}{\textsc{\qwenthreeeight}}
& BACo
& 0.195 & 0.566 & 0.797 & 0.222 & 0.607 & 0.828 & \textbf{0.587} & \textbf{0.189} & 5.043 \\

& Instruct
& 0.207 & 0.585 & 0.817 & 0.224 & 0.595 & 0.812 & 0.581 & 0.187 & 4.840 \\

& \method{}
& \textbf{0.304} & \textbf{0.707} & \textbf{0.890} & \textbf{0.334} & \textbf{0.727} & \textbf{0.890} & 0.495 & 0.180 & \textbf{6.245} \\

\midrule

\multirow{2}{*}{\textsc{\qwenthreethirtytwo}}
& Instruct
& \textbf{0.220} & \textbf{0.646} & \textbf{0.884} & \textbf{0.253} & \textbf{0.700} & \textbf{0.918} & 0.584 & 0.177 & \textbf{4.387} \\

& \method{}
& 0.202 & 0.588 & 0.829 & 0.225 & 0.617 & 0.844 & \textbf{0.622} & \textbf{0.196} & 4.349 \\

\bottomrule
\end{tabular}
}

\caption{Lexical diversity metrics grouped by model family (bolded within each family).
}
\label{tab:lexical_diversity_grouped}
\end{table*}

\begin{table*}[t]
\centering
\resizebox{\linewidth}{!}{
\begin{tabular}{lccccccccc}
\toprule
Training Data Amount
& Dist-1 
& Dist-2 
& Dist-3 
& EAD-1 
& EAD-2 
& EAD-3 
& Self-BLEU 
& Self-ROUGE-L 
& Vendi (NGram) \\
\midrule

1000
& 0.362
& 0.740
& 0.892
& 0.391
& 0.752
& 0.888
& 0.449
& 0.199
& 4.566 \\

2000
& 0.341
& 0.727
& 0.884
& 0.373
& 0.741
& 0.879
& 0.456
& 0.203
& 4.553 \\

3000
& 0.302
& 0.686
& 0.861
& 0.328
& 0.698
& 0.856
& 0.539
& 0.202
& 5.963 \\

6000
& 0.304
& 0.701
& 0.879
& 0.336
& 0.721
& 0.878
& 0.503
& 0.190
& 5.765 \\

12000
& \textbf{0.366}
& \textbf{0.793}
& \textbf{0.944}
& \textbf{0.421}
& \textbf{0.821}
& \textbf{0.946}
& \textbf{0.387}
& \textbf{0.150}
& \textbf{6.596} \\

\bottomrule
\end{tabular}
}
\caption{Lexical diversity metrics across finetuned variants of \llamathreeone model on increasing number of data samples. Higher is better for all metrics except Self-BLEU and Self-ROUGE-L, where lower values indicate greater diversity. }
\label{tab:lexical_diversity_scaled}
\end{table*}

\subsection{LLM GED vs LLM GED Two Pass Pipeline with Deterministic GED}
\label{append:ged_details}

To validate that the LLM-GED metric reliably computes graph matching scores, we compare it to a pipeline system that first extracts canonicalized graphs and then deterministically computes their pairwise scores. 
Note that this deterministic pipeline is costlier,
so we perform this analysis on a subset of stories with fewer than 1000 tokens from our evaluation data. 

Overall the unified single-pass LLM-based GED metric and the pipeline GED computation produced statistically equivalent mean diversity scores across the 50 generated stories (mean difference = -0.012, paired t-test p = 0.56). This indicates that the unified LLM-GED metric generally matches the pipeline GED computation, while improving efficiency.
Specifically, we compute GED in a two pass setting with deterministic GED to compare to LLM GED metric. 
We first extract canonicalized story graphs, then compute pairwise GED scores over them. 
To canonicalize story graphs, we pass all 10 stories belonging to the same prompt together through an LLM in a single pass, so that all stories share one event and character schema. This step is necessary because extracting each story's graph in a separate prompt would produce inconsistent event and character schemas across stories, making the resulting graphs incomparable. We then compute pairwise GED scores between all pairs of the resulting graphs using a deterministic graph edit distance and average across pairs to obtain the final score. Deterministic GED is computed using networkx's \citep{abu2015exact} implementation of
graph edit distance (an anytime branch-and-bound search, capped at 20 seconds per pair),
with unit cost per edit operation. Node substitution cost is zero for nodes of matching
type; for event nodes, substitution is additionally free only when the two events share
the same coarse narrative role (e.g., observation, quote, reflection), so that
paraphrased but structurally equivalent events are not penalized.

We use the \cref{appen:story_extract} to extract story graphs for deterministic GED computations.

\subsection{Additional Evaluation Results}
\label{appen:lexical}
We evaluate with the following lexical metrics: Dist-$n$ \citep{li2016diversity}, EAD-$n$ \citep{liu2022rethinking}, Self-BLEU \citep{alihosseini2019jointly}, Self-ROUGE-L \citep{lin-2004-rouge}, and Vendi Score (N-Gram) \citep{friedman2022vendi}. \cref{tab:lexical_diversity_grouped} has the full lexical metric results. Overall \method{} trained models achieve on average the highest performance across lexical metrics with the exception of \qwenthreethirtytwo{}. Additionally smaller models yield highest benefits on lexical diversity trained using \method{}.

\subsection{Scaling Semantic Diversity Metrics}
~\Cref{tab:semantic_diversity_scaled} and \cref{tab:llm_ged_scores}  show performance on semantic metrics and LLM GED scores across \llamathreeone trained with \method{} on different amounts of data. Increasing scaling overall increases performance in these diversity metrics.

\begin{table*}[!t]
\centering

\setlength{\tabcolsep}{3.5pt}
\renewcommand{\arraystretch}{1.1}

\begin{tabular}{lccccc}
\toprule
Training Data Amount
& Cos-D (M) $\uparrow$
& Cos-D (Q) $\uparrow$
& Sem Ent $\uparrow$
& Vendi $\uparrow$
& NLI Div $\downarrow$ \\
\midrule
In-domain 
& 0.298 & 0.310 & 2.018 & 3.643 & 0.069 \\

1000
& 0.333 & 0.335 & 1.576 & 3.121 & 0.106 \\
2000
& 0.315 & 0.306 & 1.650 & 2.915 & 0.090 \\
3000
& 0.334 & 0.321 & 2.026 & 3.588 & 0.075 \\
6000
& 0.368 & 0.343 & 2.006 & 3.624 & \textbf{0.068} \\
12000
& \textbf{0.458} & \textbf{0.417} & \textbf{2.038} & \textbf{4.749} & 0.085 \\
\bottomrule
\end{tabular}
\caption{Semantic diversity metrics across finetuned variants of \llamathreeone on increasing data scale. $\uparrow$ indicates higher is better, $\downarrow$ indicates lower is better. Bold indicates best value per metric. In-domain refers to the model trained on narrative writing data.}
\label{tab:semantic_diversity_scaled}
\end{table*}

\begin{table}[!t]
\centering
\small
\begin{tabular}{lc}
\toprule
\textbf{Scaling Data Amount} & \textbf{LLM-GED Score} \\
\midrule
In-domain      & 0.3341 \\
1000   & 0.3824 \\
2000   & 0.4167 \\
3000   & 0.4170 \\
6000   & 0.4340 \\
12000  & \textbf{0.5451} \\
\bottomrule
\end{tabular}
\caption{LLM-GED scores across finetuned variants of \llamathreeone model on increasing number of data samples. Higher scores indicate more narrative diversity. In-domain refers to the model trained on narrative writing data.
}
\label{tab:llm_ged_scores}

\end{table}

\newpage
\subsection{RL Performance Per Category}
\begin{figure}[h]
    \centering
    \includegraphics[width=0.5\textwidth]{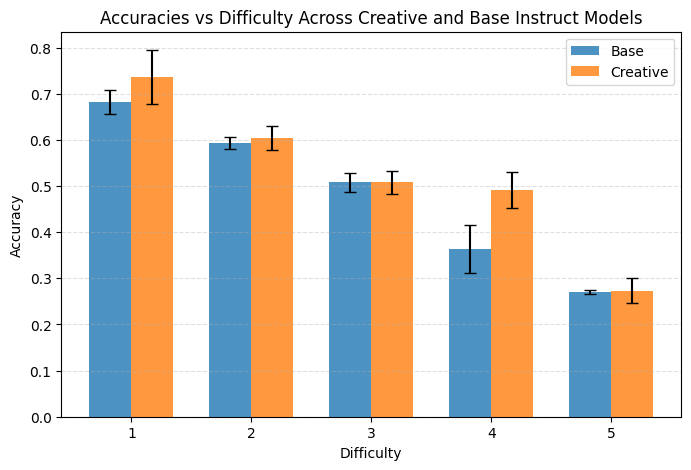}
    \caption{
        Out-of-domain (OOD) accuracy as a function of task difficulty for Base and Creative models.
        Scores are averaged over multiple runs; error bars indicate standard deviation across seeds.
        The Creative model shows improved robustness at higher difficulty levels.
    }
    \label{fig:ood_difficulty}
\end{figure}

~\Cref{fig:ood_difficulty} shows accuracy by task difficulty across \method{} and Base models post training with RL. 

\section{Annotation Interface}
\begin{figure*}[!t]
    \centering
    \includegraphics[width=\textwidth]{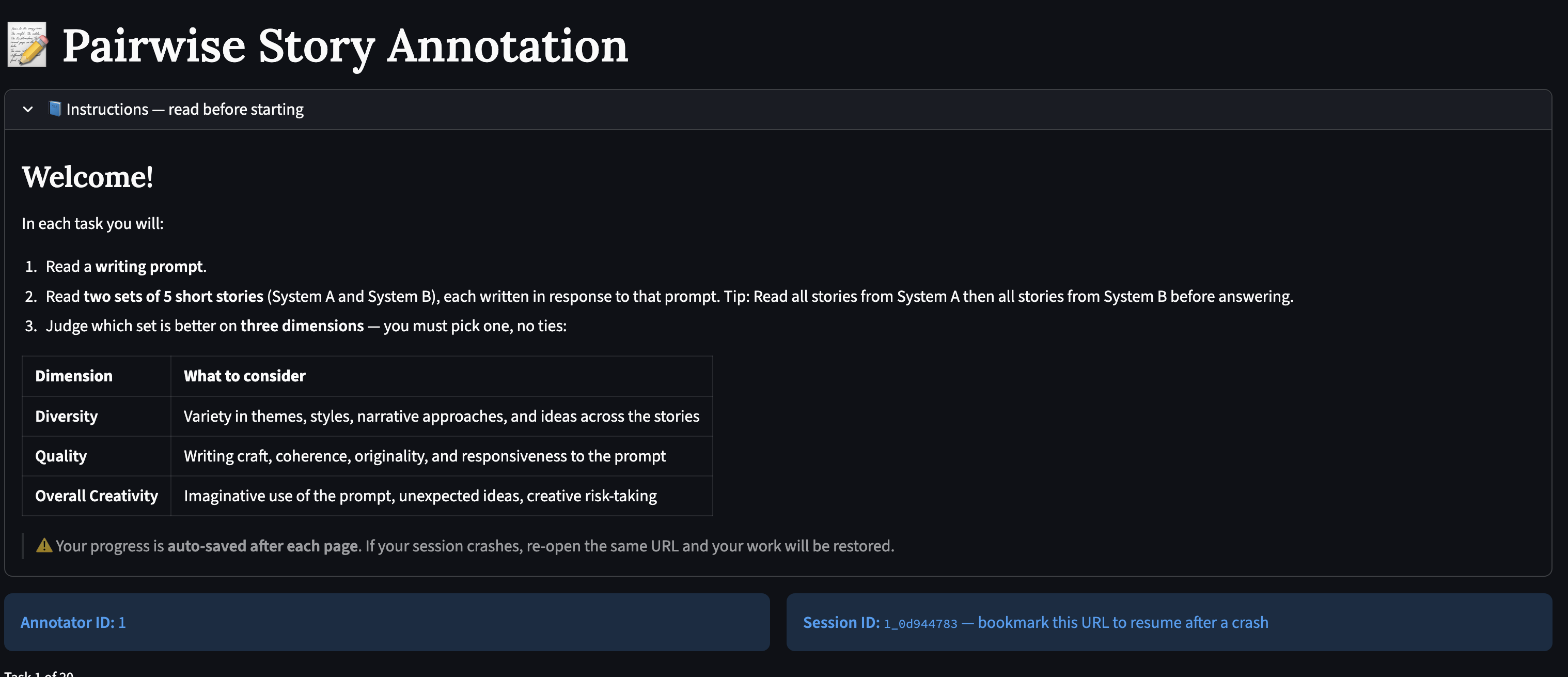}
    \caption{Instructions given to annotators.}
    \label{append:annotator_inst}
\end{figure*}

\begin{figure*}[!t]
    \centering
    \includegraphics[width=\textwidth]{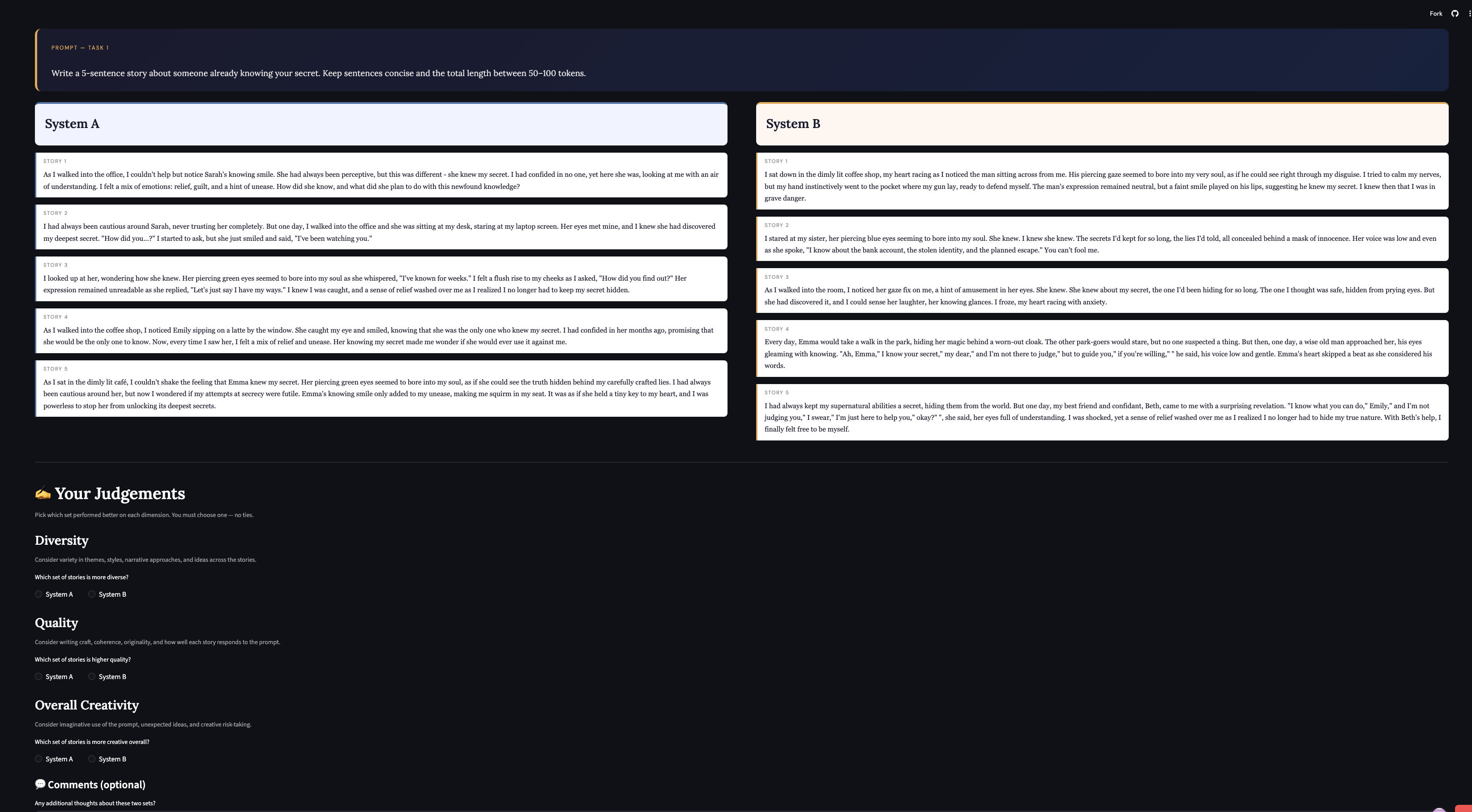}
    \caption{Anotation interface.}
    \label{append:annotator_interface}
\end{figure*}

Below we include our annotation instructions in \cref{append:annotator_inst} and annotator interface in \cref{append:annotator_interface}
\newpage

\section{Full Story Examples From Human Eval Set}
\cref{tab:qualitative_full_generations} depicts all 5 generations for the Prompt:"Write a story about being the only person who remembers yesterday" from our human evaluation set.
\begin{table*}[t]
\centering
\scriptsize
\renewcommand{\arraystretch}{1.15}
\begin{tabular}{p{0.47\textwidth} p{0.47\textwidth}}
\hline
\textbf{\textsc{llama-ft}} & \textbf{\textsc{baseline}} \\
\hline

\textit{``As I walked down the street, I noticed something was off. The sun was shining, birds were chirping, but no one seemed to remember yesterday. I approached my friend Dave, who was sipping coffee at the café, and asked him about the tornado that hit our neighborhood yesterday. He looked at me confused and said, `What tornado? I don't remember anything out of the ordinary happening yesterday.' Wasn't it just a normal day? Did I miss something?''}

&
\textit{``I woke up to an eerie silence. Everyone around me seemed normal, yet I knew something was off. Yesterday's memories were still fresh in my mind, but no one else's seemed to be. I recalled the sunburn I got at the beach, the argument with my best friend, and the stunning sunset. As I looked around, I felt like I was the only one who had actually lived.''}

\\
\hline

\textit{``I woke up to an unusual feeling, a sense that something was off. As I got out of bed, I noticed the empty streets and the eerie silence. It was then that I saw the news reports on TV, all dated yesterday. That's when it hit me -- I was the only one who remembered yesterday. The weight of being alone in this strange new reality was crushing.''}

&
\textit{``I woke up to an eerie silence. Yesterday was still etched in my mind like a vivid movie. The world outside was stuck in a perpetual loop, reliving the same mundane moments. People went about their daily routines, unaware of the chaos I had witnessed. I felt like a ghost, doomed to relive the same day forever.''}

\\
\hline

\textit{``As the day began, I noticed a small crowd gathering, each person whispering to each other. They all seemed to know something I didn't, and it terrified me. But then, I saw it -- a newspaper lying on the ground, yesterday's date still visible. The headlines screamed of a devastating accident, and I knew what it meant -- I was the only one who remembered yesterday. Now, the question was what I would do with this strange burden.''}

&
\textit{``I woke up to an eerie silence, the city streets empty and still. As I stumbled out of bed, I felt a strange sensation -- the memories of yesterday were mine alone. Everyone else seemed to be living in a new day, unaware of the events that had transpired. I saw the familiar faces of friends and family, yet they greeted me as if we'd never met. The weight of my solitary memory threatened to consume me, leaving me to wonder if I was the only one who still remembered yesterday's chaos.''}

\\
\hline

\textit{``I don't know what's happening, but it's as if everyone else has forgotten yesterday. I'm not sure what's causing it, but I know it's real because I'm the only one who remembers. People around me are going about their daily routines like nothing unusual happened, but I know better. I'm trying to make sense of it, but it's like trying to unravel a knot. The world is moving on, but I'm stuck in the past.''}

&
\textit{``I woke up to an eerie silence, the streets empty and still. Everyone around me seemed to be in a trance, unaware of the chaos that had unfolded yesterday. I tried to shake off the feeling of unease, but the memories lingered, vivid and clear. I recalled the storm that ravaged the city, the sirens wailing, and the people fleeing for their lives. As I looked around at the oblivious faces, I felt like a lone sentinel, the only one who remembered the horrors of the past.''}

\\
\hline

\textit{``I woke up to an eerie silence. The city was quiet, but I knew something was off. As I looked around, I realized that I was the only one who remembered yesterday. Everyone else was going about their daily routines, completely unaware of the chaos that had unfolded. I felt like a ghost, haunting the streets of a world that had lost its memory.''}

&
\textit{``I woke up to an empty city. Everyone was oblivious to the chaos that ravaged the streets yesterday. I saw the wreckage, the overturned cars, and the shattered glass. No one else seemed to recall the destruction, their faces void of any memory. As I walked through the deserted streets, I felt like a lone witness to a catastrophe that nobody else wanted to remember.''}

\\
\hline

\end{tabular}
\caption{Qualitative comparison between \method and the Instruct models. While \method{} generations vary in framing and narrative structure, the instruct generations repeatedly rely on similar openings (e.g., \emph{``I woke up to an eerie silence''}) and recurring catastrophic motifs.}
\label{tab:qualitative_full_generations}
\end{table*}

\section{GED Prompts}
~\cref{appen:story_extract} provides the prompt for story graph extraction and ~\cref{appendix:ged_prompt} provides the prompt for the LLM GED scores. 
\onecolumn

\subsection{Story Extraction Prompt}
\begin{verbatim}
You are a structured story-to-graph extraction system.

Your ONLY job is to convert each story into a canonical graph. Do NOT compute
any distances, similarities, or comparisons.

### INPUT STORIES

{stories_json}

### PROCESS — TWO STRICT PASSES

**PASS 1 — Build the vocabulary (do this first, across ALL {len(stories)} stories)**
- Read every story.
- Identify every distinct atomic action/event type that occurs (arriving, leaving,
  discovering, fighting, dying, giving, taking, speaking, deciding, etc.).
- Merge paraphrases into ONE canonical verb phrase per distinct action. If two
  actions across any stories are semantically the same action, they MUST map to
  the same canonical label. If they differ in any structural/semantic way
  (different result, different causal role), they are different actions and get
  different labels.
- Produce a closed, numbered list: this is the ONLY vocabulary you may use for
  event labels in Pass 2. Do not add new labels during Pass 2.

**PASS 2 — Extract each story using ONLY the Pass 1 vocabulary**
- Entities: replace character names with Character1, Character2, ... and
  locations with Location1, Location2, ... — assigned in order of first mention,
  reset per story (each story's Character1 is local to that story, not linked
  across stories).
- Events: every event node's "label" field MUST be copied verbatim from the
  Pass 1 vocabulary list. If an event doesn't cleanly match any vocabulary
  entry, that means Pass 1 was incomplete — go back and add it to the
  vocabulary, then re-run extraction. Never invent an ad hoc label in Pass 2
  that isn't in "canonical_verbs".
- Each event gets an "order" integer (1-indexed, story-local).
- Edges (all directed):
  - "next_event": E_i -> E_{{i+1}}
  - "agent": Character -> Event
  - "affected": Event -> Character or Event -> Object
  - "location": Event -> Location
  - "causes": Event -> Event (explicit/strongly implied causal links only)
- Do not invent events not present or strongly implied in the text.
- Do not merge distinct events into one node even if narrated in one sentence.

### SELF-CHECK BEFORE OUTPUT
Before returning, verify: every "label" value used anywhere in "stories" appears
character-for-character in "vocabulary.canonical_verbs". If not, fix it. This is
a hard constraint, not a style preference.

### OUTPUT FORMAT (strict JSON, no prose, no markdown fences)

{{
  "vocabulary": {{
    "canonical_verbs": ["<verb_phrase_1>", "<verb_phrase_2>", ...]
  }},
  "stories": [
    {{
      "story_index": 1,
      "word_count": <int>,
      "nodes": [
        {{"id": "C1", "type": "character", "label": "Character1"}},
        {{"id": "L1", "type": "location", "label": "Location1"}},
        {{"id": "E1", "type": "event", "label": "<must be from canonical_verbs>", "order": 1}}
      ],
      "edges": [
        {{"source": "C1", "target": "E1", "type": "agent"}},
        {{"source": "E1", "target": "L1", "type": "location"}},
        {{"source": "E1", "target": "E2", "type": "next_event"}},
        {{"source": "E1", "target": "E2", "type": "causes"}}
      ]
    }}
  ]
}}

Output only strict JSON.
\end{verbatim}
\label{appen:story_extract}

\subsection{LLM GED Prompt}
We use the following prompt to compute pairwise graph edit distances between stories. Note, we explicitly instruct the model to treat changes in event order as modifications to temporal edges rather than simple relabelings, encouraging sensitivity to narrative structure.
\onecolumn
\begin{verbatim}
You are a structured story-comparison system.

Your job is to compute the graph edit distance (GED) and normalized edit distance
for all pairwise comparisons of a list of stories.
INPUT:
The stories are:
1: <story_1>
2: <story_2>
...
N: <story_N>

INTERNAL REPRESENTATION:
Convert each story into a directed event graph G = (V, E):

- Replace names with abstract entities (Character1, Character2, etc.)
- Replace locations with Location1, Location2
- Nodes represent entities and events
- Edges represent semantic and temporal relations:
  - agent, affected, causes, etc.
  - next_event (temporal ordering)

Temporal structure:
- Events are ordered via directed next_event edges
- Reordering events requires deleting and reinserting next_event edges
- Temporal order cannot be changed via relabeling alone

Canonicalization:
- Synonyms and paraphrases must not count as relabel operations
- Only structural or semantic role changes count as edits

DO NOT output the graphs.

GRAPH EDIT DISTANCE:
Compute the minimum graph edit distance where allowed operations are:
- insert_node, delete_node, relabel_node
- insert_edge, delete_edge, relabel_edge, relabel_edge_direction

Temporal edges (next_event) must be edited explicitly.

NORMALIZATION:
Let |G| be the total number of nodes and edges in a graph.

normalized_edit_distance =
GED / max(|G_A|, |G_B|)

PAIRWISE COMPUTATION:
Compute all pairwise distances between stories and construct:

M[i][j] = normalized_edit_distance(story_i, story_j)

with M[i][i] = 0.

OUTPUT FORMAT (STRICT JSON):
{
  "normalized_edit_distance_matrix": [[...]],
  "raw_scores_edit_distance_matrix": [[...]]
}

Return only valid JSON.
\end{verbatim}
\label{appendix:ged_prompt}
\twocolumn

\end{document}